\documentclass[11pt,a4paper]{article}
\usepackage[hyperref]{acl2021}
\usepackage{times}
\usepackage{latexsym}

\usepackage{amsmath,amsfonts,bm}

\def\eqref#1{equation~\ref{#1}}
\def\1{\bm{1}}

\def\va{{\bm{a}}}

\def\vx{{\bm{x}}}

\def\mE{{\bm{E}}}

\def\mW{{\bm{W}}}

\DeclareMathAlphabet{\mathsfit}{\encodingdefault}{\sfdefault}{m}{sl}
\SetMathAlphabet{\mathsfit}{bold}{\encodingdefault}{\sfdefault}{bx}{n}

\def\gG{{\mathcal{G}}}

\def\sE{{\mathbb{E}}}

\def\sV{{\mathbb{V}}}

\newcommand{\softmax}{\mathrm{softmax}}

\usepackage{subcaption}
\usepackage{gensymb}
\usepackage{multirow}
\usepackage{graphicx}
\usepackage{float}
\usepackage{paralist}

\usepackage{microtype}

\aclfinalcopy % Uncomment this line for the final submission
\title{Generating Landmark Navigation Instructions from Maps as a Graph-to-Text Problem}

\author{Raphael Schumann \\
  Computational Linguistics\\
  Heidelberg University, Germany \\
  \And
  Stefan Riezler \\
  Computational Linguistics \& IWR \\
  Heidelberg University, Germany \\
  \hspace{-7cm}\texttt{\{rschuman|riezler\}@cl.uni-heidelberg.de}}

\date{}

\begin{document}
\maketitle
\begin{abstract}
Car-focused navigation services are based on turns and distances of named streets, whereas navigation instructions naturally used by humans are centered around physical objects called landmarks. We present a neural model that takes OpenStreetMap representations as input and learns to generate navigation instructions that contain visible and salient landmarks from human natural language instructions. Routes on the map are encoded in a location- and rotation-invariant graph representation that is decoded into natural language instructions. Our work is based on a novel dataset of 7,672 crowd-sourced instances that have been verified by human navigation in Street View. Our evaluation shows that the navigation instructions generated by our system have similar properties as human-generated instructions, and lead to successful human navigation in Street View.
\end{abstract}

\section{Introduction}
\begin{figure*}
  \centering
  \begin{subfigure}{0.99\textwidth}
  \includegraphics[width=\textwidth]{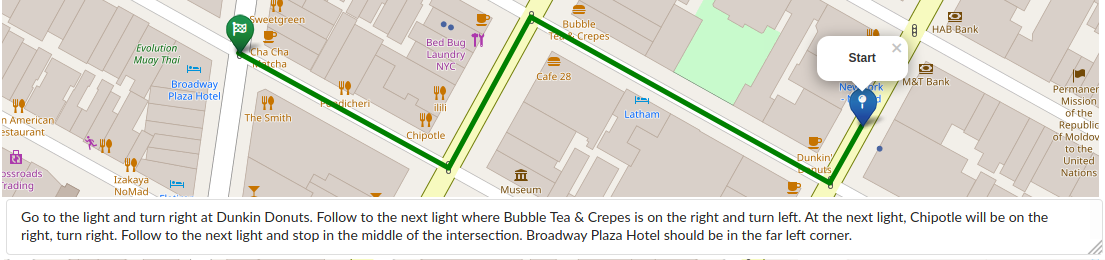}
  \vspace{0.01cm}
  \end{subfigure}
  \begin{subfigure}{0.99\textwidth}
  \includegraphics[width=\textwidth]{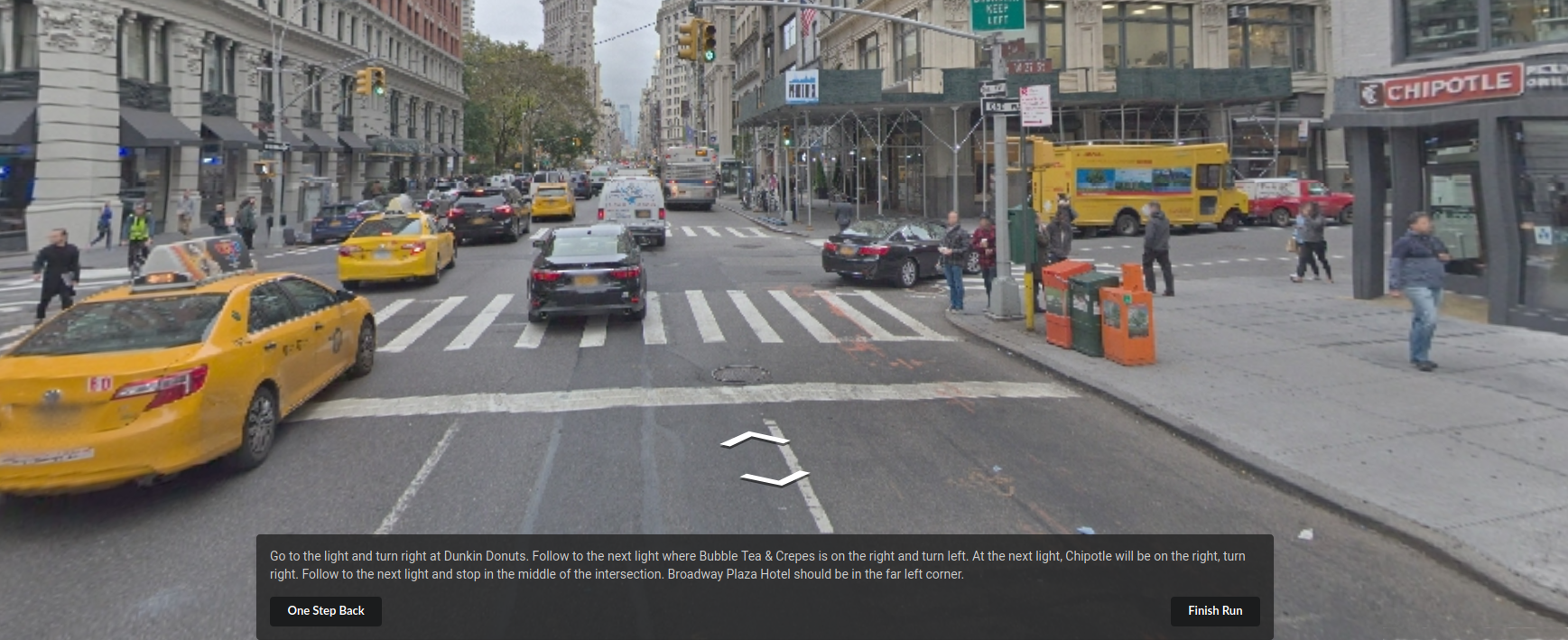}
  \end{subfigure}
  \caption{The data collection is split into two tasks. In the navigation instructions task (top) annotators see a rendered map and write instructions to follow the route. The navigation run task (bottom) is used to validate navigation instructions. A different annotator tries to find the goal location in Street View.}
  \label{fig:instructions_task}
\end{figure*}
Current navigation services provided by the automotive industry or by Google Maps generate route instructions based on turns and distances of named streets. In contrast, humans naturally use an efficient mode of navigation based on visible and salient physical objects called landmarks. As shown by \citet{tom_and_denis_2004}, route instructions based on landmarks are easier processed and memorized by humans. \citet{May2003} recommend that in pedestrian navigation systems, "landmarks should be used as the primary means of providing directions". Another navigation scenario where landmarks are useful is if GPS tracking is poor or not available, and if information is inexact regarding distances (e.g., in human estimates) or street names (e.g., for users riding a bicycle). We present a neural model that takes a real-world map representation from OpenStreetMap\footnote{\url{www.openstreetmap.org}} as input and generates navigation instructions that contain salient landmarks, learned directly from human natural language instructions.

In our framework, routes on the map are learned by discretizing the street layout, connecting street segments with adjacent points of interest, thus encoding visibility of landmarks, and encoding the route and surrounding landmarks in a location- and rotation-invariant graph. Based on crowd-sourced natural language instructions for such map representations, a graph-to-text mapping is learned that decodes graph representations into natural language route instructions that contain salient landmarks.
Our work is accompanied by a dataset of 7,672 instances of routes in OpenStreetMap and corresponding crowd-sourced natural language instructions. The navigation instructions were generated by workers on the basis of maps including all points of interest, but no street names. They were verified by different workers who followed the navigation instructions on Google Street View\footnote{\url{www.google.com/streetview}}.

Experimental results on randomly sampled test routes show that our graph-to-text model produces landmarks with the same frequency found in human reference instructions. Furthermore, the time-normalized success rate of human workers finding the correct goal location on Street View is 0.664. Since these routes can have a partial overlap with routes in the training set, we further performed an evaluation on completely unseen routes. The rate of produced landmarks drops slightly compared to human references, and the time-normalized success rate also drops slightly to 0.629. While there is still room for improvement, our results showcase a promising direction of research, with a wide potential of applications in various existing map applications and navigation systems.

The main contributions of this paper are:
\begin{compactitem}[$\bullet$]
\item We collect and publish a large scale dataset of natural language landmark navigation instructions that are validated by human navigation runs in Street View.
\item We present a method to represent geospatial routes as a graph and propose an appropriate graph-to-text architecture that learns to generate navigation instructions from real-world data.
\end{compactitem}

\section{Related Work and Datasets}

\citet{mirowski-etal-2018-learning} published a subset of Street View covering parts of New York City and Pittsburgh. Street View is a navigable environment that is build from connected real-world 360\degree panoramas. This data is used by \citet{hermann-etal-2019-learning} to train a visual agent to follow turn-by-turn instructions generated by Google Maps API. 
\citet{chen-etal-2019-touchdown} published a Street View dataset\footnote{\url{www.streetlearn.cc}} with more recent and higher resolution panorama images that covers the lower half of Manhattan. They further introduce the Touchdown task that has the goal to navigate Street View in order to find a hidden teddy bear. The data for that task is obtained from annotation workers that follow a predefined route in Street View and write down navigation instructions along the way. 
A central difference between Touchdown and our dataset is the annotation modality: Touchdown annotators use  panorama images along the route, while our instruction writers only see the rendered route on a map. See Section~\ref{sec:dataset} for a more detailed discussion.

Our work puts the task of natural language navigation upside down by learning to generate human-like navigation instructions from real-world map data instead of training an agent to follow human generated instructions. Prior work in this area has used rule-based systems to identify landmarks \citep{rousell-zipf-2017-towards} or to generate landmark-based navigation instructions \citep{drager-koller-2012-generation,cercas-curry-etal-2015-generating}. Despite having all points of interest on the map available, our approach learns to verbalize only those points of interest that have been deemed salient by inclusion in a human navigation instruction. Previous approaches that learn navigation instructions from data have been confined to simplified grid-based representations of maps for restricted indoor environments \citep{daniele-etal-2017-navigational}.
\citet{devries-etal-2018-talk} tackles the problem in a more sophisticated outdoor environment but the model fails to verbalize useful instructions when conditioned on more than one possible landmark. Other work generates navigation instructions from indoor panoramas along a path but provides no explicit evaluation like human navigation success. They rather use the instructions to augment the training routes for a vision and language navigation agent \citep{NEURIPS2018_6a81681a}.

\begin{table*}[ht!]
\resizebox{.99\textwidth}{!}{
\begin{tabular}{|l|r|ll|rr|rr|}
\hline
\textbf{Dataset} & \multicolumn{1}{c|}{\textbf{\#Instructions}} & \textbf{Environment} & \textbf{Data Source} & \multicolumn{1}{c}{\textbf{\#Nodes}} & \multicolumn{1}{c|}{\textbf{Avg. Length}} & \multicolumn{1}{c}{\textbf{Vocabulary}} & \multicolumn{1}{c|}{\textbf{Avg. Tokens}} \\ \hline
Talk the Walk    & 786                                          & gridworld            & 3D rendering         & 100                                  & 6.8                                       & 587                                     & 34.5                                      \\
Room-to-Room     & 21,567                                       & indoor               & panoramas            & 10,800                               & 6.0                                       & 3,156                                   & 29.0                                      \\
Touchdown        & 9,326                                        & outdoor              & panoramas            & 29,641                               & 35.2                                      & 4,999                                   & 89.6                                      \\
Talk2Nav         & 10,714                                       & outdoor              & panoramas and map    & 21,233                               & 40.0                                      & 5,240                                   & 68.8                                      \\
Room-X-Room      & 126,069                                      & indoor               & panoramas            & 10,800                               & 7.0                                       & 388K                                    & 78.0                                      \\
\textbf{map2seq} & \textbf{7,672}                               & \textbf{outdoor}     & \textbf{map}         & \textbf{29,641}                      & \textbf{40.0}                             & \textbf{3,826}                          & \textbf{55.1}                             \\ \hline
\end{tabular}}
\caption{Overview of natural language navigation instructions datasets. The instructions in our dataset rely solely on information present in OpenStreetMap. \textbf{Dataset}: Talk the Walk~\citep{walkthetalk}; Room-to-Room~\citep{room2room}; Touchdown~\citep{chen-etal-2019-touchdown}; Talk2Nav~\citep{vasudevan2020talk2nav}; Room-X-Room \citep{rxr}; map2seq (this work). \textbf{\#Instructions}: Number of instructions in the dataset. \textbf{Environment}: Type of navigation environment. \textbf{Data Source}: Type of information the annotator uses to write the navigation instructions. \textbf{\#Nodes}: Number of nodes in the discretized environment. \textbf{Avg. Length}: Average number of nodes per route. \textbf{Vocabulary}: Number of unique tokens in the instructions. \textbf{Avg. Tokens}: Number of tokens per route instruction.}
\label{tab:datasets}
\end{table*}
\section{Task}

The task addressed in our work is that of automatically generating Natural Language Landmark Navigation Instructions (NLLNI) from real-world open-source geographical data from OpenStreetMap. The instructions are generated \textit{a priori}~\citep{janarthanam-etal-2012-web} for the whole route. Training data for NLLNI was generated by human crowdsourcing workers who were given a route on an OpenStreetMap rendering of lower Manhattan, with the goal of producing a succinct natural language instruction that does not use street names or exact distances, but rather is based on landmarks. Landmarks had to be visible on the map and included, e.g., churches, cinemas, banks, shops, and public amenities such as parks or parking lots. Each generated navigation instruction was validated by another human crowdsourcing worker who had to reach the goal location by following the instruction on Google Street View. 

NLLNI outputs are distinctively different from navigation instructions produced by OpenRouteService, Google Maps, or car navigation systems. While these systems rely on stable GPS signals such that the current location along a grid of streets can be tracked exactly, we aim at use cases where GPS tracking is not available, and knowledge of distances or street names is inexact, for example, pedestrians, cyclists, or users of public transportation. The mode of NLLNI is modeled after human navigation instructions that are naturally based on a small number of distinctive and visible landmarks in order to be memorizable while still being informative enough to reach the goal.  
A further advantage of NLLNI is that they are based on map inputs which are more widely available and less time dependent than Street View images.

\begin{table*}
\centering
\resizebox{.94\textwidth}{!}{
\begin{tabular}{|l|cc|cc|cc|l|}
\hline
\multicolumn{1}{|c|}{\multirow{2}{*}{\textbf{Phenomenon}}} & \multicolumn{2}{c|}{\textbf{R-to-R}} & \multicolumn{2}{c|}{\textbf{Touchdown}} & \multicolumn{2}{c|}{\textbf{map2seq}} & \multicolumn{1}{c|}{\multirow{2}{*}{\textbf{Example}}}                         \\ \cline{2-7}
\multicolumn{1}{|c|}{}                                     & $c$              & $\mu$             & $c$               & $\mu$               & $c$              & $\mu$              & \multicolumn{1}{c|}{}                                                          \\ \hline
Reference to unique entity                                 & 25               & 3.7               & 25                & 9.2                 & 25               & 6.3                & ... turn right where \textbf{Dough Boys} is on the corner ...                  \\
Coreference                                                & 8                & 0.5               & 15                & 1.1                 & 8                & 0.5                & ... is a bar, Landmark tavern, stop outside of \textbf{it} ...                 \\
Comparison                                                 & 1                & 0.0               & 3                 & 0.1                 & 0                & 0.0                & ... there are two lefts, \textbf{take the one that is not sharp} ...           \\
Sequencing                                                 & 4                & 0.2               & 21                & 1.6                 & 24               & 1.8                & ... continue straight at the \textbf{next} intersection ...                    \\
Count                                                      & 4                & 0.2               & 9                 & 0.4                 & 11               & 0.6                & ... go through the next \textbf{two} lights ...                                \\
Allocentric spatial relation                               & 5                & 0.2               & 17                & 1.2                 & 9                & 0.5                & ... go through the next \textbf{light with Citibank at the corner.} ...        \\
Egocentric spatial relation                                & 20               & 1.2               & 23                & 3.6                 & 25               & 3.2                & ... at the end of the park \textbf{on your right}...                           \\
Imperative                                                 & 25               & 4.0               & 25                & 5.2                 & 25               & 5.3                & ... \textbf{head} down the block and \textbf{go} through the double lights ... \\
Direction                                                  & 22               & 2.8               & 24                & 3.7                 & 25               & 3.5                & ... head \textbf{straight} to the light and make a \textbf{right} ...          \\
Temporal condition                                         & 7                & 0.4               & 21                & 1.9                 & 7                & 0.3                & ... go straight \textbf{until you come} to the end of a garden area ...        \\
State verification                                         & 2                & 0.1               & 18                & 1.5                 & 12               & 0.6                & ... \textbf{you should see} bike rentals on your right ...                     \\ \hline
\end{tabular}}
\caption{Linguistic analysis of 25 randomly sampled navigation instructions. Numbers for Room-to-Room~\citep{room2room} and Touchdown~\citep{chen-etal-2019-touchdown} taken from the latter. $c$ is the number of instructions out of the $25$ which contain the phenomenon at least once. $\mu$ is the mean number of times each phenomenon occurs.}
\label{tab:linguistics}
\end{table*}
\section{Data Collection}
Because there is no large scale dataset for NLLNI that is generated from map information only, we collect data via crowdsourcing. The annotator is shown a route on the map and writes navigation instructions based on that information~(Figure~\ref{fig:instructions_task},~top). We take the approach of \citet{chen-etal-2019-touchdown} and determine correctness of navigation instructions by showing them to other annotators that try to reach the goal location in Street View~(Figure~\ref{fig:instructions_task},~bottom).

\subsection{Resources and Preparation}
\label{sec:res_and_prep}
We use the static Street View dataset provided by \citet{chen-etal-2019-touchdown}. This allows us to make the experiments in this work replicable. Because the panorama pictures were taken at the end of 2017, we export an OpenStreetMap extract of Manhattan from that time. OpenStreetMap (OSM) is an open source collection of geodata that can be used to render maps of the world. It features detailed street layouts and annotations for points of interest (POI) like amenities, infrastructure or land use\footnote{\url{openstreetmap.org/wiki/Map_Features}}.
We discretize the street layout by creating a node every ten meters along the roads. The resulting structure is further referenced to as the OSM graph with nodes consisting of street segments. Based on that graph, we sample routes of length between 35 and 45 nodes. A route is the shortest path between its start and end node. It includes a minimum of three intersections (i.e., a node with more than two edges) and ends in proximity to a POI. We further assure that it is possible to follow the route in Street View by verifying that a corresponding subgraph exists in the Street View graph.

\subsection{Crowdsourcing}
We use Amazon Mechanical Turk (AMT)\footnote{\url{www.mturk.com}} to acquire annotators. Before working on the actual tasks, workers were required to pass a tutorial and qualification test. The tutorial introduces the tasks, teaches basic mechanics of Street View and explains meaning of map icons. A feature of AMT and additional IP address\footnote{IP addresses were not saved and are not part of the dataset.} lookup ensures that annotators are located in the United States. This increases the probability of working with native English speakers and people familiar with US street environments. We paid \$0.35 per navigation instructions task and \$0.20 for the navigation run task. Furthermore, we paid a bonus of \$0.15 for successfully reaching the goal location and \$0.25 for validated navigation instructions. The amounts were chosen on the basis of \$10/hour.
The annotation procedure involved two phases. First, an annotator wrote navigation instructions for a given route. Afterwards, a different annotator used the instructions to navigate to the goal location. If one of two annotators did so successfully, the navigation instructions were considered valid.

\paragraph{Navigation Instructions Task}
As shown in Figure~\ref{fig:instructions_task}~(top), the annotator sees a route on a map which is rendered without street names. Workers were told to write navigation instructions as if "a tourist is asking for directions in a neighborhood you are familiar with" and to "mention landmarks to support orientation". The navigation instructions were written in a text box below the map which is limited to 330 characters.

\paragraph{Navigation Run Task}
Figure~\ref{fig:instructions_task}~(bottom) shows the Street View interface with navigation instructions faded-in at the bottom. It is possible to look around 360\degree~and movement is controlled by the white arrows. In addition there is a button on the bottom left to backtrack which proved to be very helpful. The initial position is the start of the route facing in the correct direction. The annotators finish the navigation run with the bottom right button either when they think the goal location is reached or if they are lost. The task is successful if the annotator stops the run within a 25 meter radius around the goal location.

\begin{figure*}[ht!]
    \centering
    \includegraphics[width=0.99\textwidth]{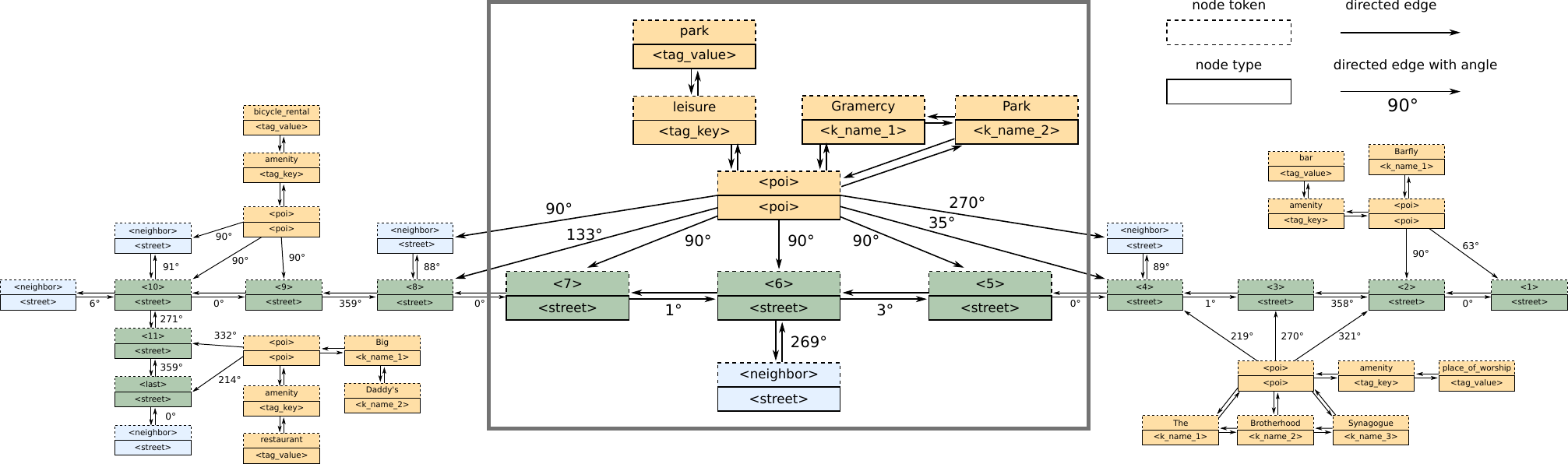}
    \caption{Graph representation of the route in Figure~\ref{fig:graph_construction}. The framed middle part is magnified for readability. Some nodes are left out for sake of clear visualization. Also, node colors are for visualization only and not encoded in the graph. Green nodes are part of the route. Blue nodes are neighboring street segments. Orange nodes belong to OSM points of interest. Angles are relative to route direction and start clockwise at 0\degree~which is facing forward.}
    \label{fig:graph}
\end{figure*}

\begin{figure}[ht!]
  \includegraphics[width=0.49\textwidth]{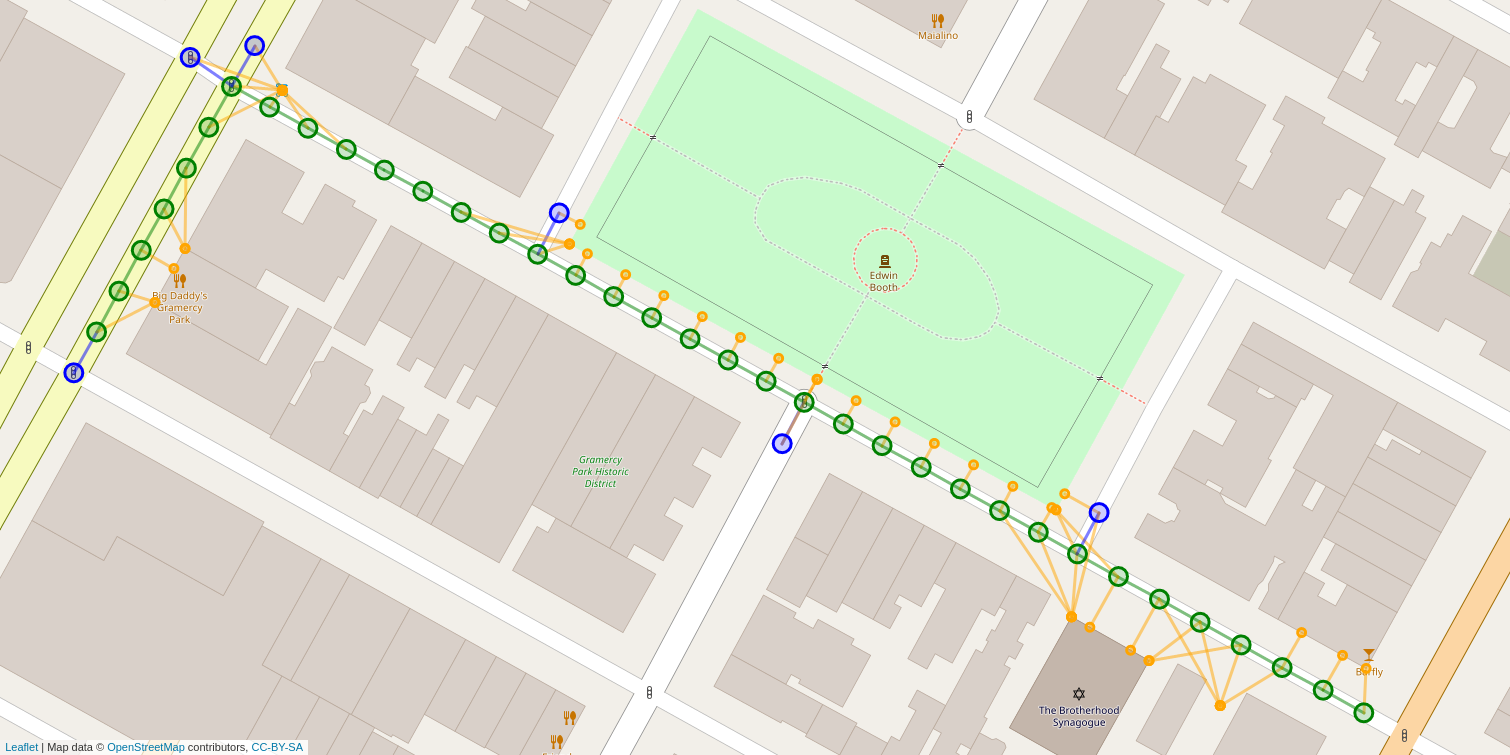}
  \caption{Route rendered on the map with street segments and landmark visibility.}
  \label{fig:graph_construction}
\end{figure}
\subsection{Dataset}
\label{sec:dataset}
The data collection resulted in \textbf{7,672 navigation instructions that were manually validated in Street~View}. For additional 1,059 instructions, the validation failed, which amounts to a validation rate of 88\%. Of the validated instructions, 1,033 required a second try in the navigation run task. On average, instructions are 257 characters long, with a minimum length of 110, and a maximum of 330 characters. We release the segmented OSM graph, the routes in that graph paired with the collected navigation instructions, and the data split used in our experiments\footnote{\url{www.cl.uni-heidelberg.de/statnlpgroup/map2seq/}}.
Table~\ref{tab:datasets} gives a comparison of different datasets with natural language landmark navigation instructions. Our dataset is the only one that uses only map information to generate navigation instructions. The advantage of relying solely on map data is the global availability and longevity of the encoded features. In contrast, navigation instructions written from Street View include temporary features like construction utilities, street advertisements, or passing vehicles.
Table~\ref{tab:linguistics} shows a qualitative linguistic analysis of the navigation instructions of different datasets. In general, navigation instructions are driven by giving directions in imperative formulation while referencing to entities along the route. In contrast to the Touchdown task where including store names was prohibited, the entities in our instructions are often referenced to by their name. Although the instruction writers in our setting did not see the route in first person perspective, objects are vastly referenced to in egocentric manner (egocentric with respect to the navigating agent). This is because the annotator knows the starting direction and can infer the facing direction for the rest of the route. Because the initial facing direction in Touchdown is random, the first part of their instructions is about rotating the agent. This explains the higher number of occurrences of the state verification phenomenon. In our dataset, state verification is usually used to ensure the correct stopping position. The different setting of data collection is also reflected by the temporal condition phenomenon. Annotators of Touchdown write down instructions while navigating Street View and thus experience the temporal component first hand, while our annotators have a time independent look at the route. 

\section{Method}

The underlying OSM geodata of the rendered map is an XML tree of nodes located in the latitude-longitude coordinate system. The nodes are composed into ways and polygons\footnote{\url{www.openstreetmap.org/wiki/Elements}}. These elements in connection with their annotations are used to render the visual map.
In the next subsection we propose our approach to represent a route and its surrounding map features as a graph that includes all necessary information for generating landmark navigation instructions. The second subsection describes the neural graph-to-text architecture that is trained to learn inductive representations of the individual route graphs and to decode navigation instructions from them.

\subsection{Map-to-Graph Representation}

The basis of the graph for a single route is the OSM subgraph~(Section \ref{sec:res_and_prep}) that includes the actual route nodes. Further, neighboring street segment nodes are added. This is depicted in Figure~\ref{fig:graph_construction} as green and blue circles, respectively. In order to decide on the visibility of the POIs, we employ a technique similar to that of \citet{rousell-zipf-2017-towards}. For each street segment, the POIs in a radius of 30 meters are identified. If a line drawn between the street segment and the POI is not interrupted by a building polygon, the POI is considered visible from that particular street segment. If the POI itself is (inside) a polygon, then the line is drawn to the closest point on the POI polygon. The orange circles in Figure~\ref{fig:graph_construction} show the results of the visibility check and how they naturally fit into the graph structure. Each point of interest in OSM has one or more tags in the form of key and value pairs. They store properties like type or name.
Note that we only determine the geometric visibility of the POIs and do not incorporate any hand-crafted salience scores as to what would be a good landmark. Instead, saliency of a landmark is implicitly learned from natural language verbalization of the POI in the human-generated instruction. 

An example graph representation of the route in Figure~\ref{fig:graph_construction} is given in Figure~\ref{fig:graph}. Formally, a route representation is a directed graph $\gG=(\sV, \sE)$, where $\sV$ denotes the set of nodes and $\sE$ the set of edges. A node $v$ consists of a node type $v^t$ and a node token $v^w$. There are $V^t$ node types and $V^w$ node tokens. Street segments are of type \textit{\textless street\textgreater}. A point of interest has the node type \textit{\textless poi\textgreater}. An OSM tag key has the node type \textit{\textless tag\_key\textgreater} and an OSM tag value has the node type \textit{\textless tag\_value\textgreater}. The node token further specifies nodes in the graph. Street segments that belong to the route have a node token \textit{\textless P\textgreater} according to their sequential position P. The last route segment has the special token \textit{\textless last\textgreater}. Other street segment nodes have the \textit{\textless neighbor\textgreater} token. The actual key and value literals of an OSM tag are the node tokens of the respective node. The OSM name tag is split into multiple nodes with type \textit{\textless k\_name\_N\textgreater} where N is the word position and the node token is the word at that position.

All adjacent street segment nodes are connected with an edge in both directions. If a POI is visible from a particular street segment, there is an edge from the corresponding POI node to that street segment node. Each POI node is connected with their tag key nodes. A tag value node is connected to its corresponding tag key node. The name tag nodes of the same POI are connected with each other. Some edges have a geometric interpretation. This is true for edges connecting a street segment with either a POI or with another street segment. These edges $(u,v) \in \sE^A, \sE^A \subset \sE$ have a label attached. The label $ang(u,v)$ is the binned angle between the nodes relative to route direction. The continuous angle $[0\degree, 360\degree)$ is assigned to one of 12 bins. Each bin covers 30\degree~with the first bin starting at 345\degree.
The geometric distance between nodes is not modeled explicitly because street segments are equidistant and POI visibility is determined with a maximum distance.
The proposed representation of a route and its surroundings as a directed graph with partially geometric edges is location- and rotation-invariant, which greatly benefits generalization. 

\subsection{Graph-to-Text Architecture}
By representing a route as a graph, we can frame the generation of NLLNI from maps as a graph-to-text problem. The encoder learns a neural representation of the input graph and the sequence decoder generates the corresponding text. The architecture follows the Transformer~\citep{vaswani-etal-2017-attention} but uses graph attentional layers~\citep{velickovic-etal-2018-graph} in the encoder. Graph attention injects the graph structure by masking (multi-head) self-attention to only attend to nodes that are first-order neighbors in the input graph. The geometric relations between some nodes are treated as edge labels which are modeled by distinct feature transformation matrices during node aggregation~\citep{gcnr}.

The input to a layer of the encoder is a set of node representations, ${\bf x} = \{\vx_1, \vx_2, \dots, \vx_N\}, \vx_i \in \mathbb{R}^{d_m}$, where $N$ is the number of nodes and $d_m$ is the model size. Each layer $l: \mathbb{R}^{d_m} \rightarrow \mathbb{R}^{d_m}$ takes ${\bf x}$ and produces new node representations ${\bf x'}$. The input to the first layer is constructed from the concatenation of type and token embedding: $\vx_i=ReLU(\mW^F[\mE^T_{v_i^t} || \mE^W_{v_i^w}])$ where $\mW^F \in \mathbb{R}^{2d_m \times d_m}$  is a weight matrix, $\mE^T \in \mathbb{R}^{d_m}$ and $\mE^W \in \mathbb{R}^{d_m}$ are embedding matrices for node types and node tokens, respectively.

The output of a single graph attention head is the weighted sum of neighboring node representations:

\begin{equation}
	\bar{\vx}_i = \sum_{j | (v_j, v_i) \in \sE}\alpha_{ij}(\mW_{r(i, j)}^U\vx_j)
\end{equation}
The weight coefficient is computed as $\alpha_{ij} = \softmax_j{(e_{ij})} = \frac{ \exp{(e_{ij})} }{ \sum_{k | (v_k, v_i) \in \sE} \exp{(e_{ik})} }$ where $e_{ij}$ measures the compatibility of two node representations:
\vspace{-0.1cm}
\begin{equation}
\resizebox{.42\textwidth}{!}{
$e_{ij} = LeakyReLU(\va^T[\mW^V \vx_i|| \mW_{r(i, j)}^U \vx_j])$
}
\end{equation}
where $\va \in \mathbb{R}^{2d_h}$, $\mW^V \in \mathbb{R}^{d_m \times d_h}$, $d_h = d_m / h$ is the attention head dimension and $h$ is the number of heads. In the case of a geometric relation between nodes, the weight matrix $\mW_{r(i, j)}^U \in \mathbb{R}^{d_m \times d_h}$ is selected according to the angle label between the nodes: $r(i, j) = ang(u_i, u_j)$, otherwise $r(i, j) = unlabeled$. The output of each head is concatenated and after a skip connection forwarded to the next encoder layer. The encoder layer is applied $L$~times and the final node representations ${\bf x^*}$ are used in the decoder context attention mechanism. Thus, no modification of the Transformer decoder is necessary and $L$~decoder layers are used. Further, the decoder can copy node tokens from the input into the output sequence~\citep{copy_point}.

The described architecture is able to model all aspects of the input graph. Graph attention models directed edges. Edge labels model the geometric relation between nodes. Heterogeneous nodes are represented by their type embedding and token embedding. The sequentiality of the route is encoded by tokens (\textit{\textless 1\textgreater}, \textit{\textless 2\textgreater}, ...) of the respective nodes. This is analogous to absolute position embeddings which provide word order information for text encoding~\citep{vaswani-etal-2017-attention, devlin-etal-2019-bert}.

\begin{table}[t]
\resizebox{.49\textwidth}{!}{
\begin{tabular}{l|cccccc|}
\cline{2-7}
                                   & \textbf{BLEU} & \textbf{Len.} & \textbf{Landm.} & \textbf{SDTW} & \textbf{SR} & \textbf{SNT}  \\ \cline{2-7} 
                                   & \multicolumn{6}{c|}{200 instances test set}                                     \\ \hline
\multicolumn{1}{|l|}{reference}    & -             & 53.5   & 2.76               & .728          & .855        & \textbf{.878} \\ \hline
\multicolumn{1}{|l|}{rule based}   & 0.71          & 53.1   & 12.44              & .405          & .460        & \textbf{.455} \\
\multicolumn{1}{|l|}{seq2seq}      & 13.12         & 52.9   & 1.95               & .139          & .160        & \textbf{.206} \\ \hline
\multicolumn{1}{|l|}{graph2text}   & 18.60         & 52.6   & 2.41               & .475          & .540        & \textbf{.676} \\
\multicolumn{1}{|l|}{g2t+pretrain} & 18.81         & 52.5   & 2.44               & .471          & .540        & \textbf{.537} \\ \hline
                                   & \multicolumn{6}{c|}{700 instances test set}                                     \\ \hline
\multicolumn{1}{|l|}{reference}    & -             & 53.5   & 2.72               & .726          & .861        & \textbf{.830} \\ \hline
\multicolumn{1}{|l|}{g2t+pretrain} & 17.39         & 53.0   & 2.41               & .475          & .551        & \textbf{.664} \\ \hline
\end{tabular}}
\caption{Evaluation of navigation instructions produced by models and human reference on \textbf{partially seen} test routes. Evaluation metrics are explained in Section \ref{sec:eval_metrics}.}
\label{tab:results_seen}
\end{table}
\begin{table}[t]
\resizebox{.49\textwidth}{!}{
\begin{tabular}{l|cccccc|}
\cline{2-7}
                                   & \textbf{BLEU} & \textbf{Len.} & \textbf{Landm.} & \multicolumn{1}{l}{\textbf{SDTW}} & \textbf{SR} & \textbf{SNT}  \\ \cline{2-7} 
                                   & \multicolumn{6}{c|}{\textit{200 instances test set}}                                                 \\ \hline
\multicolumn{1}{|l|}{reference}    & -             & 57.5   & 2.68               & .725                              & .824        & \textbf{.791} \\ \hline
\multicolumn{1}{|l|}{rule based}   & 0.67          & 52.3   & 10.96              & .472                              & .525        & \textbf{.512} \\
\multicolumn{1}{|l|}{seq2seq}      & 11.12         & 51.8   & 1.58               & .074                              & .100        & \textbf{.137} \\ \hline
\multicolumn{1}{|l|}{graph2text}   & 14.07         & 50.5   & 1.74               & .344                              & .400        & \textbf{.534} \\
\multicolumn{1}{|l|}{g2t+pretrain} & 15.64         & 50.3   & 2.33               & .367                              & .429        & \textbf{.530} \\ \hline
                                   & \multicolumn{6}{c|}{\textit{700 instances test set}}                                                 \\ \hline
\multicolumn{1}{|l|}{reference}    & -             & 54.2   & 2.69               & .727                              & .843        & \textbf{.807} \\ \hline
\multicolumn{1}{|l|}{g2t+pretrain} & 16.27         & 53.2   & 2.30               & .407                              & .473        & \textbf{.629} \\ \hline
\end{tabular}}
\caption{Evaluation of navigation instructions produced by models and human reference on \textbf{unseen} test routes.}
\label{tab:results_unseen}
\end{table}
\begin{figure*}[ht]
    \includegraphics[width=0.99\textwidth]{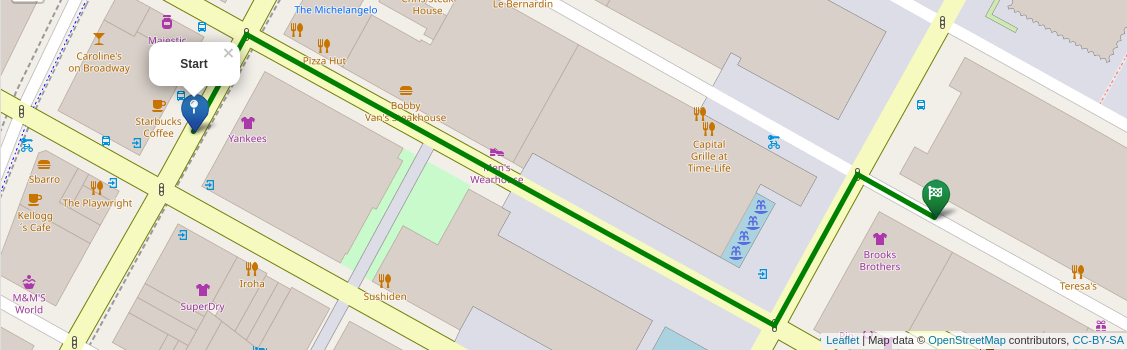}
    \scriptsize\textbf{reference:} \textit{At the light with Fridays on the corner, turn right. Continue down the long street to the next light with Nine West on the right corner, then turn left. Go to the next light with Brooks Brothers on the right corner, then turn right and stop.} \\
    \textbf{rule based:} \textit{Starbucks Coffee left subway entrance right Best Buy Mobile left Yankees right bus stop left bus stop left light right The Michelangelo left TGI Fridays left Pizza Hut left Bobby Van 's left park right Men 's Wearhouse left fountain left fountain left subway entrance left light left Nine West right Rockefeller Center left subway entrance right Brooks Brothers right light right stop} \\
    \textbf{seq2seq:} \textit{Go straight to the light and make a left. Go straight to the next light and make a left. Go straight to the light and make a right. Stop one step after turning with Brooks Brothers to your right.} \\
    \textbf{graph2text:} \textit{Walk to the light with TGI Fridays on the corner and turn right. Walk down the long block to the next light with Nine West on the left corner, then turn left. Walk to the next light with Brooks Brothers on the far right corner, then turn right.} \\
    \textbf{g2t+pretrain:} \textit{Turn right at the first set of lights with TGI Fridays on the left corner. Pass a park on the right and turn left at the lights. Pass the fountain on the right and turn right at the lights. Take two steps and stop. Brooks Brothers is on the right corner.}
    \caption{Route from partially seen test set paired with instructions generated by different systems.}
    \label{fig:example}
\end{figure*}
\section{Experiments}

\subsection{Baselines}
We consider two baselines. A \textbf{rule based} system that uses a single heuristic to construct instructions by stringing together all POIs and intersections along the route, and following each intersection by the turning direction. Similar, POIs are followed by 'left' or 'right' depending on which side of the street they appear. The end of the route is signaled by the 'stop' token.  The second baseline is a \textbf{seq2seq}~(sequence-to-sequence)~model that is trained on pairs of rule based navigation instructions and crowdsourced instructions. The seq2seq model follows the Transformer architecture~\citep{vaswani-etal-2017-attention} with copy mechanism and is trained with the same hyperparameters as the graph-to-text model. Examples are given in Figure~\ref{fig:example}.

\subsection{Experimental Setup}
We construct a graph for each route as described above. On average there are 144 nodes in a graph and 3.4 edges per node. There are 8 different node types and a vocabulary of 3,791 node tokens. The hyperparameters for the graph-to-text architecture are set as follows: The embedding and hidden size is set to 256. We use 6 encoder and decoder layers with 8 attention heads. Cross entropy loss is optimized by Adam \citep{adam} with a learning rate of 0.5 and batch size of 12. The embedding matrix for node tokens and output tokens is shared. Additionally we experiment with pretraining the graph-to-text model with above mentioned rule based instructions as target. This teaches the model sequentiality of route nodes and basic interpretation of the angle labels. We generate 20k instances for pretraining and further fine tune on the human generated instances. Both models and the seq2seq baseline are trained on 5,667 instances of our dataset. The best weights for each model are selected by token accuracy based early stopping on the 605 development instances.

\subsection{Evaluation Metrics}
\label{sec:eval_metrics}

\textbf{BLEU} is calculated with SacreBLEU~\citep{sacrebleu} on lower-cased and tokenized text.

\noindent\textbf{Length} is the average length in number of tokens.

\noindent\textbf{Landmarks} is the number of landmark occurrences per instance. Occurrences are identified by token overlap between navigation text and tag values of POIs along the route. E.g., landmarks in the instructions in Figure~\ref{fig:instructions_task} are: \textit{Dunkin Donuts, Bubble Tea \& Crepes, Chipotle, Broadway Hotel}.

\noindent\textbf{SDTW} is success weighted by normalized Dynamic~Time~Warping~\citep{ndtw}. Distance between two nodes is defined as meters along the shortest path between the two nodes and threshold distance is 25 meters.

\noindent\textbf{SR} is the first try success rate in the navigation run task. Success is achieved if the human navigator stops within a radius of 25 meters around the goal.

\noindent\textbf{SNT} is success weighted by navigation time: $\frac{1}{N}\sum_{i=1}^{N}{S_i\frac{\bar{t}_i}{t_i}}$, where $S_i$ is a binary success indicator that is 1 if the annotator stops within a 25 meter radius around the goal. $t_i$~is the time until the navigation run is finished. We empirically estimate the expected navigation time $\bar{t}_i$ as 1.3 seconds\footnote{Average over all successful navigation runs in the dataset.} per node in the route. This estimation ranges from 45.5 seconds for routes with 35 nodes to 58.5 seconds for routes with 45 nodes. SNT is inspired by SPL~\citep{SPL} but considers trajectory time instead of trajectory length.

\subsection{Experimental Results and Analysis}

Results of our experimental evaluation are shown in Table~\ref{tab:results_seen}~and~\ref{tab:results_unseen}. We evaluate on unseen data, i.e., routes without any overlap with routes in the training set, and on partially seen data, i.e., routes randomly sampled from the training area with partial overlaps.\footnote{The data split is shown in the Appendix.} For the baseline models we perform the human evaluation on a 200 instances subset of the full 700 instances test set.

On the partially seen test set with 200 instances, our proposed graph-to-text models outperform the baseline models in terms of the success based metrics. In the unseen setup, the rule based baseline achieves a better success rate, but falls short when success is weighted by navigation time. This result shows that the instructions generated by the rule based system are exact by including all possible landmarks, but obviously do not resemble natural language and high evaluation time suggests that they are hard to read. Despite moderate BLEU scores and reasonable amount of produced landmarks, the seq2seq baseline fails to generate useful navigation instructions. The pretrained graph-to-text model performs better than its plain counterpart in the unseen setup. It produces more correct landmarks and higher success rates. In the extended evaluation the pretrained graph-to-text model is compared with the reference on 700 instances in each test set. Under the central evaluation metric of success normalized by time (SNT), our model reaches .664 and .629 on partially seen and unseen test data, respectively.

An example output for each system together with the input map is shown in Figure~\ref{fig:example}. The rule based instruction is complete, but ignores saliency of landmarks and is hard to read. The seq2seq baseline generates a navigation instruction that sounds human-like and also includes salient landmarks found on the map. However, the directions are incorrect in this example. The graph-to-text based models get the directions right and produce fluent natural language sentences. They include landmarks at the correct sequential position. A further qualitative evaluation of instructions generated by the graph-to-text models is given in the Appendix.

\section{Conclusion}
We presented a dataset and suitable graph-to-text architecture to generate landmark navigation instructions in natural language from OpenStreetMap geographical data. Our neural model includes novel aspects such as a graphical representation of a route using angle labels. Our dataset consists of a few thousand navigation instructions that are verified for successful human navigation. The dataset is large enough to train a neural model to produce navigation instructions that are very similar in several aspects to human-generated instructions on partially seen test data. However, performance naturally drops on unseen data including new types of landmarks in new combinations.

\subsubsection*{Acknowledgments}
We would like to thank Christian Buck and Massimiliano Ciaramita for initial fruitful discussions about this work.
The research reported in this paper was supported by a Google Focused Research Award on "Learning to Negotiate Answers in Multi-Pass Semantic Parsing". 

\bibliographystyle{acl_natbib}
\bibliography{acl2021}

\clearpage
\appendix
\addcontentsline{toc}{section}{Appendices}
\section*{Appendices}
\section{Dataset Split}
\begin{figure}[ht]
    \centering
    \includegraphics[width=0.4\textwidth]{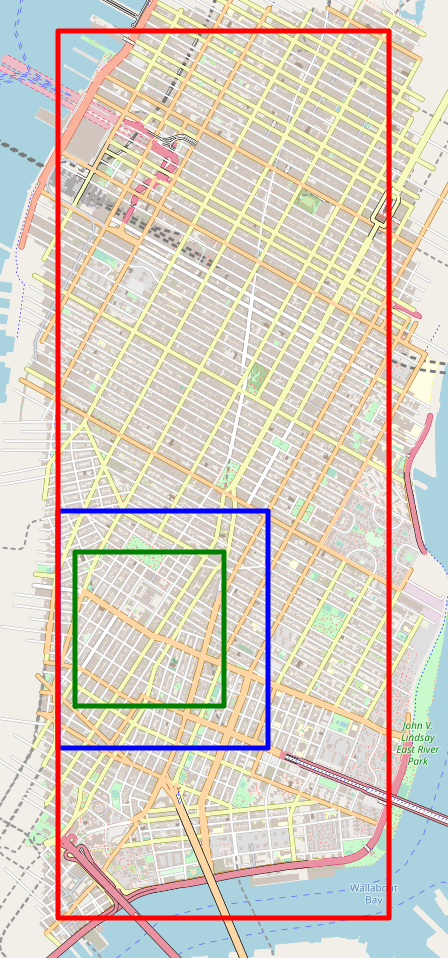}
    \caption{Dataset splits}
    \label{fig:splits}
\end{figure}
All 700 routes that are exclusively in the green rectangle are in the unseen test set. All 605 routes that cross the green border are in the development set. None of those development set routes extend further than the blue rectangle. The training set consists of routes within the red rectangle but outside of the green rectangle. The partially seen test set consists of 700 randomly sampled routes from the training set (and removed from the training set). Partially seen means that subsequences of those routes can be present in the training set.

\section{Evaluation Navigation Success Rate Analysis}
\begin{figure}
  \centering
  \begin{subfigure}{0.49\textwidth}
  \includegraphics[width=\textwidth]{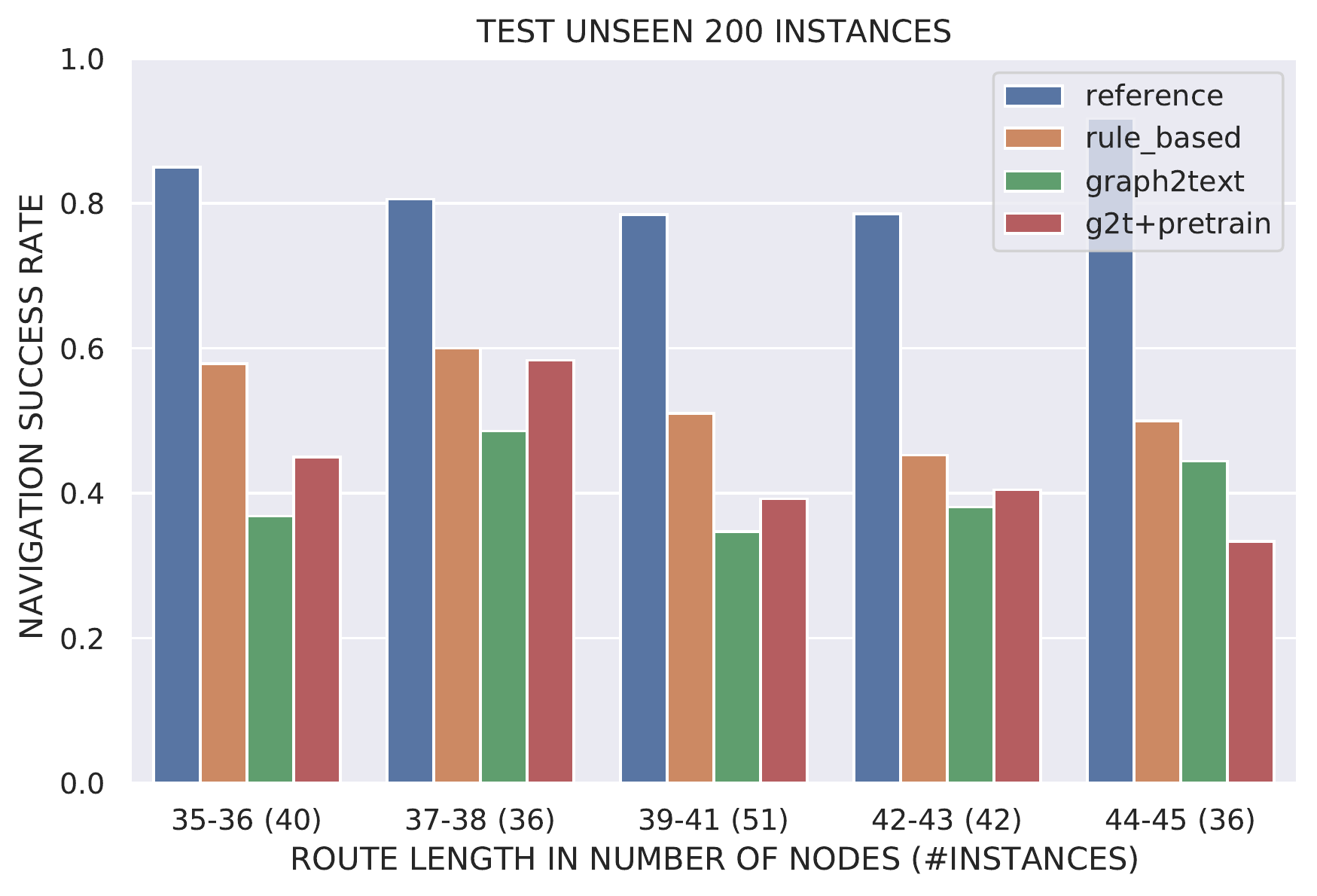}
  \end{subfigure}
  \begin{subfigure}{0.49\textwidth}
  \includegraphics[width=\textwidth]{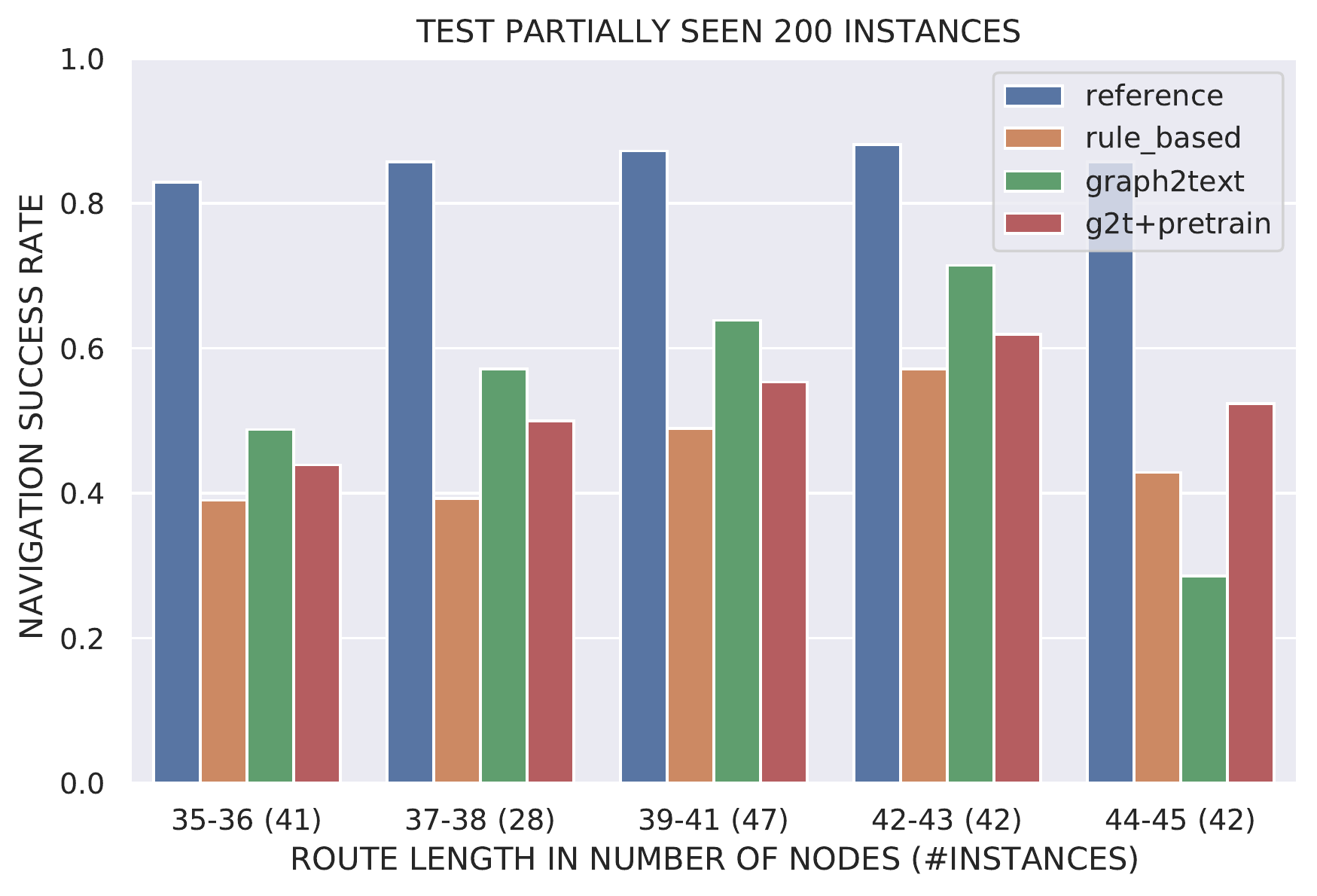}
  \end{subfigure}
  \caption{Navigation success rate in respect of route length. Length is measured in number of nodes in a route.}
  \label{fig:length_success}
\end{figure}

\begin{figure}
  \centering
  \begin{subfigure}{0.48\textwidth}
  \includegraphics[width=\textwidth]{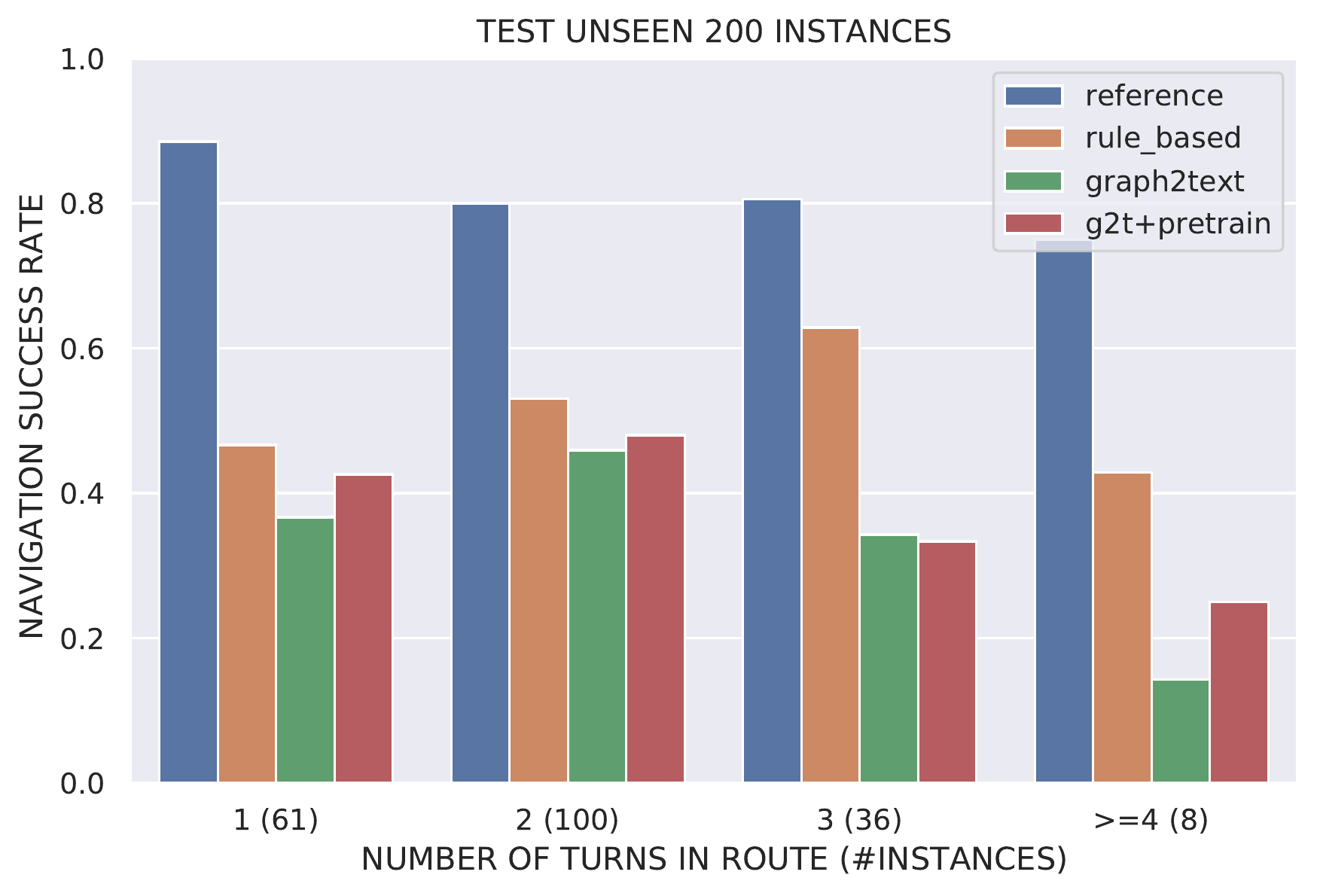}
  \end{subfigure}
  \begin{subfigure}{0.48\textwidth}
  \includegraphics[width=\textwidth]{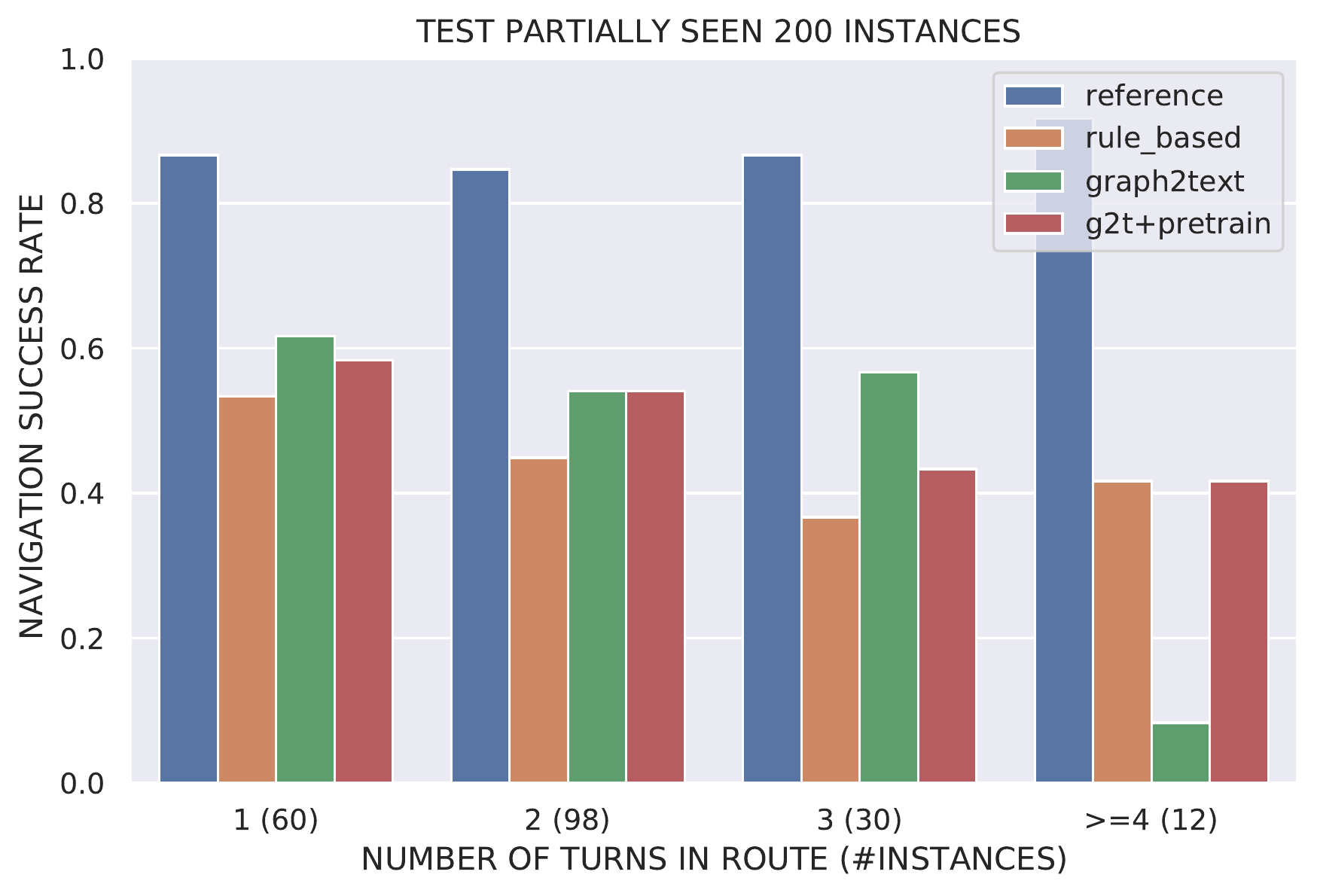}
  \end{subfigure}
  \caption{Navigation success rate in respect of number of turns in a route. A turn is defined as an intersection that isn't crossed in straight direction (345\degree to 15\degree).}
  \label{fig:turns_success}
\end{figure}

We analyze the navigation success rate with respect to properties of the corresponding routes. Figure~\ref{fig:length_success} shows that the length of the route has little influence on the navigation success rate on the partially seen test set. On the unseen data there is tendency in favor of shorter routes for the g2t+pretrain model. The reference instructions do not show such bias. Figure~\ref{fig:turns_success} shows navigation success with respect to number of turns in a route which is another complexity indicator. The success rate drops with an increasing number of turns for all systems but not for the reference instructions. The analysis reveals that performance of our model drops with increasing route complexity while it is stable for reference instructions. The rule based system appears to be more stable with increasing number of turns in comparison to the learned models.
\begin{figure}
  \centering
  \begin{subfigure}{0.48\textwidth}
  \includegraphics[width=\textwidth]{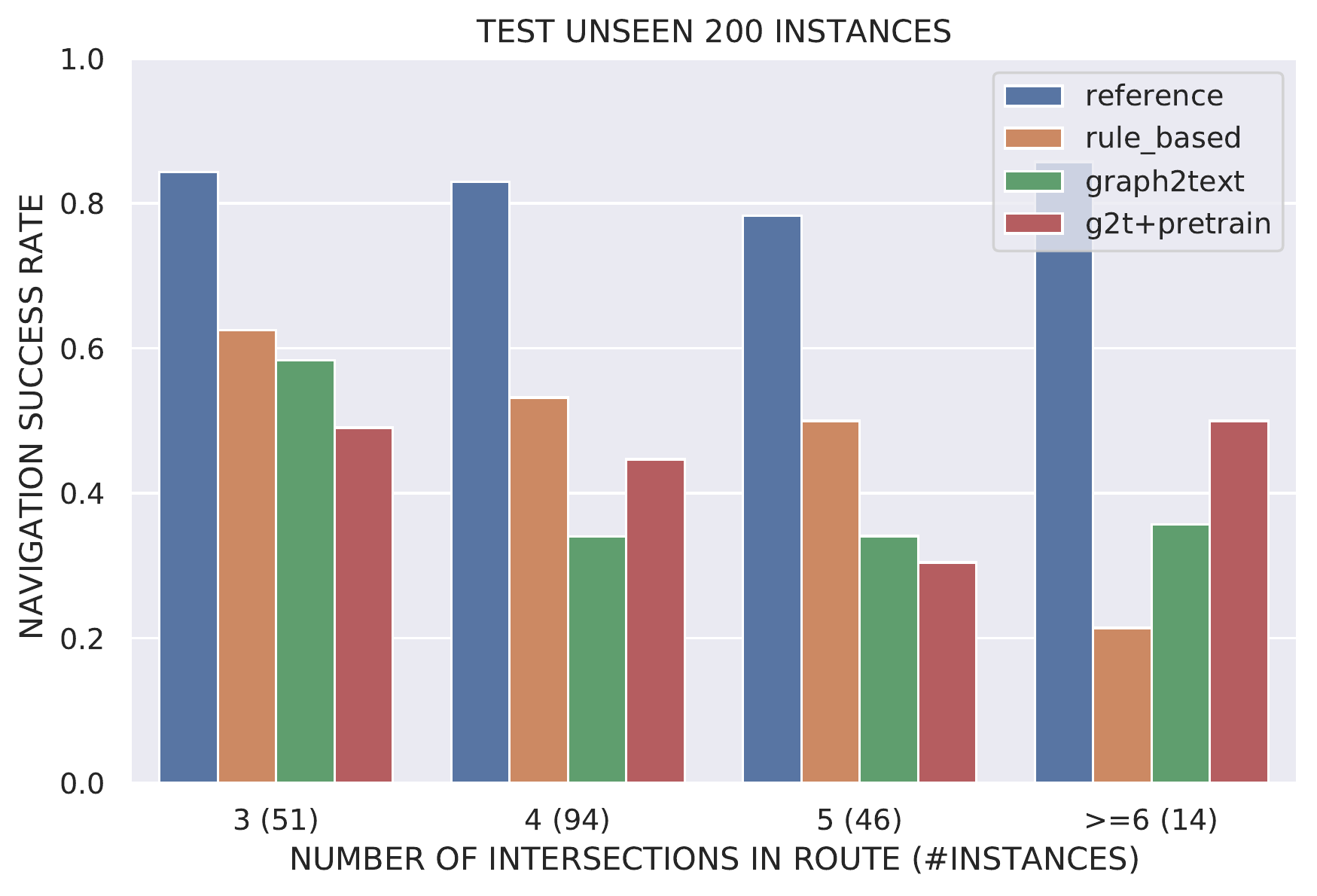}
  \end{subfigure}
  \begin{subfigure}{0.48\textwidth}
  \includegraphics[width=\textwidth]{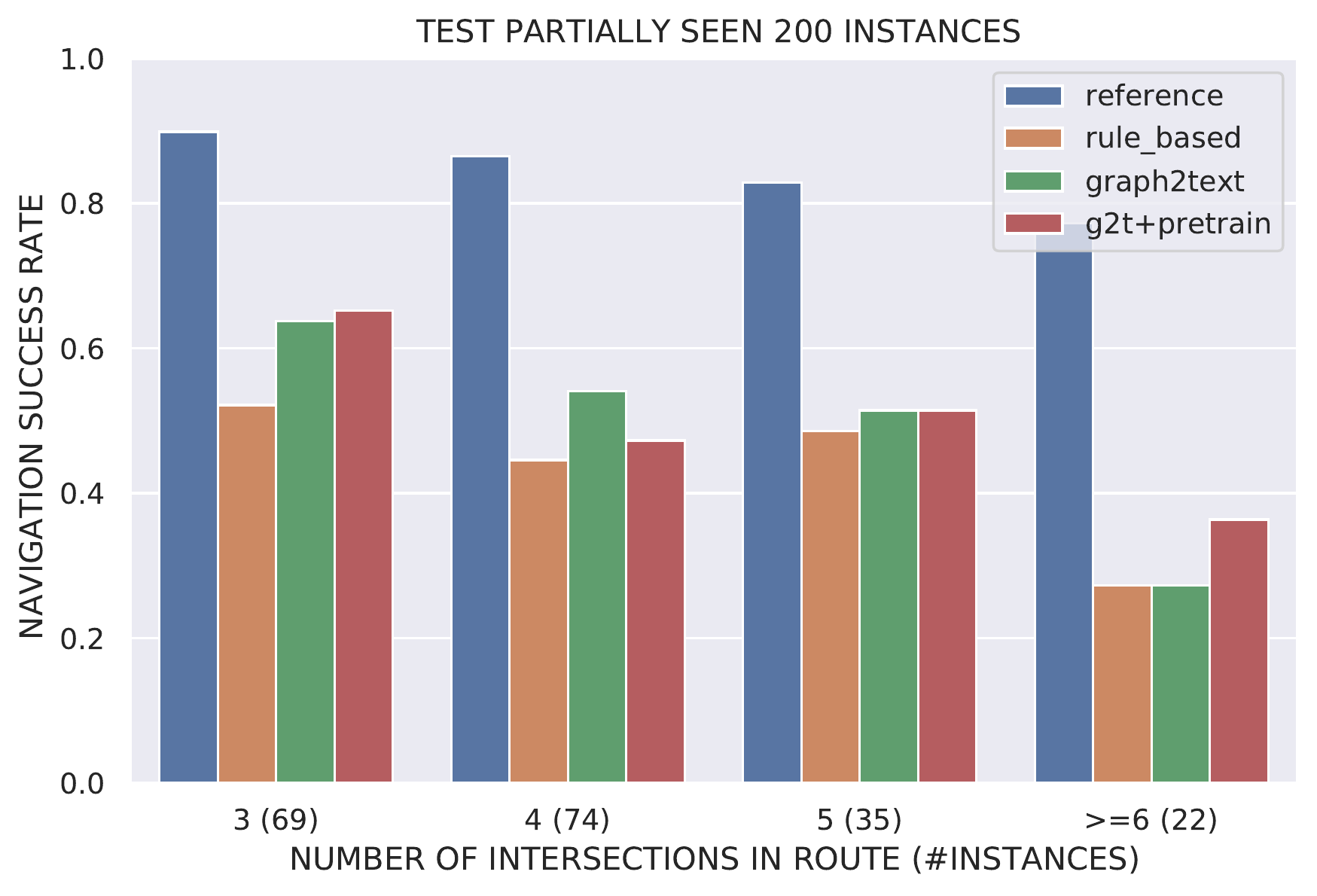}
  \end{subfigure}
  \caption{Navigation success rate in respect of number of intersections in a route. Each node in the route with more than two neighbors is counted as an intersection.}
  \label{fig:intersections_success}
\end{figure}

\section{Landmarks}
Table~\ref{tab:top_landmarks_seen}~and~\ref{tab:top_landmarks_unseen} presents a scoring of types of landmarks produced by our pretrained model. A comparison of landmarks produced in human-generated reference instructions to those produced in model-generated instructions shows a large overlap on partially seen data, and ranking is similar to hand-crafted salient scores used in work in geoinformatics \citep{rousell-zipf-2017-towards}. The distribution of landmarks in the unseen test data is different from the partially seen data. To some extent, the model is able to adapt to the unseen environment.
\begin{table}
\resizebox{.49\textwidth}{!}{
\begin{tabular}{l|lll|lll|}
\cline{2-7}
\multicolumn{1}{c|}{\textbf{}}     & \multicolumn{3}{c|}{\textbf{Reference}}                                    & \multicolumn{3}{c|}{\textbf{Model}}                                        \\ \hline
\multicolumn{1}{|c|}{\textbf{Top}} & \multicolumn{2}{c}{\textbf{OSM tag}} & \multicolumn{1}{c|}{\textbf{Score}} & \multicolumn{2}{c}{\textbf{OSM tag}} & \multicolumn{1}{c|}{\textbf{Score}} \\ \hline
\multicolumn{1}{|l|}{1}            & amenity:     & bank                  & 0.41                                & amenity:        & pharmacy           & 0.39                                \\
\multicolumn{1}{|l|}{2}            & leisure:     & park                  & 0.35                                & shop:           & furniture          & 0.38                                \\
\multicolumn{1}{|l|}{3}            & amenity:     & pharmacy              & 0.32                                & amenity:        & bank               & 0.37                                \\
\multicolumn{1}{|l|}{4}            & shop:        & furniture             & 0.30                                & leisure:        & garden             & 0.29                                \\
\multicolumn{1}{|l|}{5}            & cuisine:     & burger                & 0.29                                & cuisine:        & burger             & 0.28                                \\
\multicolumn{1}{|l|}{6}            & leisure:     & garden                & 0.29                                & shop:           & supermarket        & 0.25                                \\
\multicolumn{1}{|l|}{7}            & cuisine:     & coffee\_shop          & 0.26                                & cuisine:        & coffee\_shop       & 0.25                                \\
\multicolumn{1}{|l|}{8}            & amenity:     & place\_of\_worship    & 0.25                                & cuisine:        & american           & 0.24                                \\
\multicolumn{1}{|l|}{9}            & cuisine:     & american              & 0.23                                & shop:           & convenience        & 0.22                                \\
\multicolumn{1}{|l|}{10}           & amenity:     & bicycle\_rental       & 0.23                                & cuisine:        & italian            & 0.21                                \\ \hline
\end{tabular}}
\caption{Frequency of OSM tags of landmark occurrences in the instructions for the \textbf{partially seen test set}, normalized by the number of occurrences in the input graph.}
\label{tab:top_landmarks_seen}
\end{table}
\begin{table}
\resizebox{.49\textwidth}{!}{
\begin{tabular}{l|lll|lll|}
\cline{2-7}
\multicolumn{1}{c|}{\textbf{}}     & \multicolumn{3}{c|}{\textbf{Reference}}                                    & \multicolumn{3}{c|}{\textbf{Model}}                                        \\ \hline
\multicolumn{1}{|c|}{\textbf{Top}} & \multicolumn{2}{c}{\textbf{OSM tag}} & \multicolumn{1}{c|}{\textbf{Score}} & \multicolumn{2}{c}{\textbf{OSM tag}} & \multicolumn{1}{c|}{\textbf{Score}} \\ \hline
\multicolumn{1}{|l|}{1}            & amenity:        & cinema             & 0.58                                & cuisine:        & juice              & 0.64                                \\
\multicolumn{1}{|l|}{2}            & shop:           & wine               & 0.53                                & amenity:        & pharmacy           & 0.55                                \\
\multicolumn{1}{|l|}{3}            & shop:           & computer           & 0.53                                & shop:           & convenience        & 0.50                                \\
\multicolumn{1}{|l|}{4}            & amenity:        & pharmacy           & 0.51                                & amenity:        & cinema             & 0.46                                \\
\multicolumn{1}{|l|}{5}            & cuisine:        & coffee\_shop       & 0.49                                & cuisine:        & coffee\_shop       & 0.46                                \\
\multicolumn{1}{|l|}{6}            & tourism:        & hotel              & 0.44                                & shop:           & computer           & 0.45                                \\
\multicolumn{1}{|l|}{7}            & shop:           & convenience        & 0.42                                & tourism:        & hotel              & 0.41                                \\
\multicolumn{1}{|l|}{8}            & shop:           & houseware          & 0.31                                & shop:           & pet                & 0.39                                \\
\multicolumn{1}{|l|}{9}            & shop:           & supermarket        & 0.31                                & shop:           & beauty             & 0.38                                \\
\multicolumn{1}{|l|}{10}           & amenity:        & bank               & 0.28                                & shop:           & wine               & 0.38                                \\ \hline
\end{tabular}}
\caption{Frequency of OSM tags of landmark occurrences in the instructions for the \textbf{unseen test set}, normalized by the number of occurrences in the input graph.}
\label{tab:top_landmarks_unseen}
\end{table}

\section{Annotation Instructions}
The AMT workers got the following instructions for the writing task:

The goal of this task is to write navigation instructions for a given route. Imagine a tourist is asking for directions in a neighborhood you are familiar with and try to mention useful landmarks to support orientation. Another annotator will later read your instructions in order to find the goal location in StreetView (Navigation Run Task). If the other annotator successfully navigates to the goal location, your instruction is validated.

\section{Examples}

\begin{figure*}[ht]
    \includegraphics[width=0.99\textwidth]{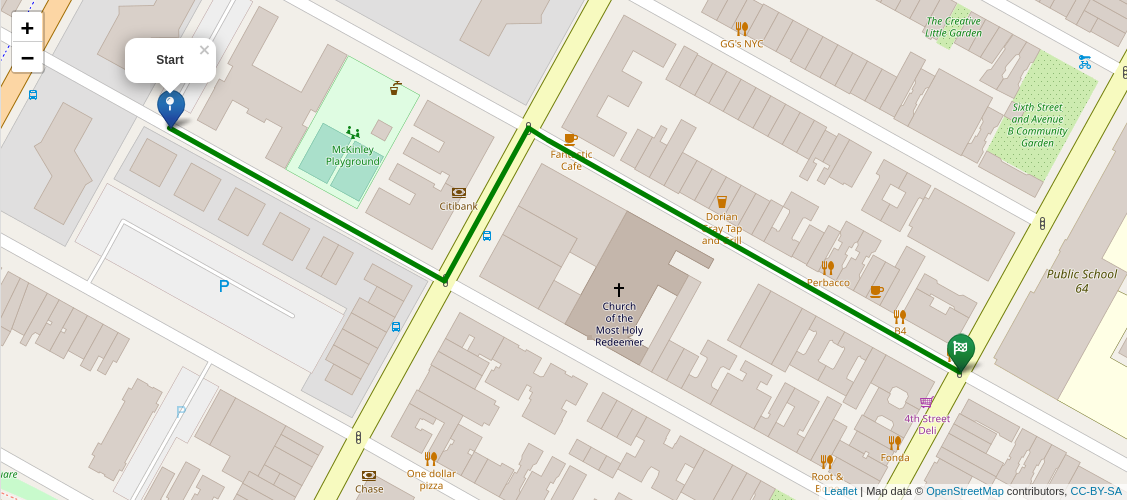}
    \scriptsize
    \textbf{reference:} \textit{Head straight past the playground and turn left at the light. Turn right at the next light with Fantastic Cafe on the right. Continue down the very long street and stop at the next light with turntable 5060 on the near left corner.} \\
    \textbf{rule based:} \textit{parking left pitch left playground left pitch left bus stop right light left Citibank left light right Fantastic Cafe left Church of the right Dorian Gray Tap left Perbacco left Matcha Cafe Wabi left B4 left stop}\\
    \textbf{seq2seq:} \textit{Go straight to the light and turn left. Go straight to the first light and turn left. Go straight to the first light and stop a couple of steps before the next light. You should see Perbacco on your left.}\\
    \textbf{graph2text:} \textit{Head to the light and turn left. At the next light with Fantastic cafe on your near right corner, turn right. Go to the next light and stop in the middle of the intersection. B4 Gray Tap should be on your left.}\\
    \textbf{g2t+pretrain:} \textit{Head to the light and turn left. At the following light with Fantastic cafe on the far left corner, turn right. Go to the next light with 4th street Deli on the near left corner and stop in the middle of the intersection.}
    \caption{Route from partially seen test set with successful navigation for g2t+pretrain.}
\end{figure*}

\begin{figure*}[h]
    \includegraphics[width=0.99\textwidth]{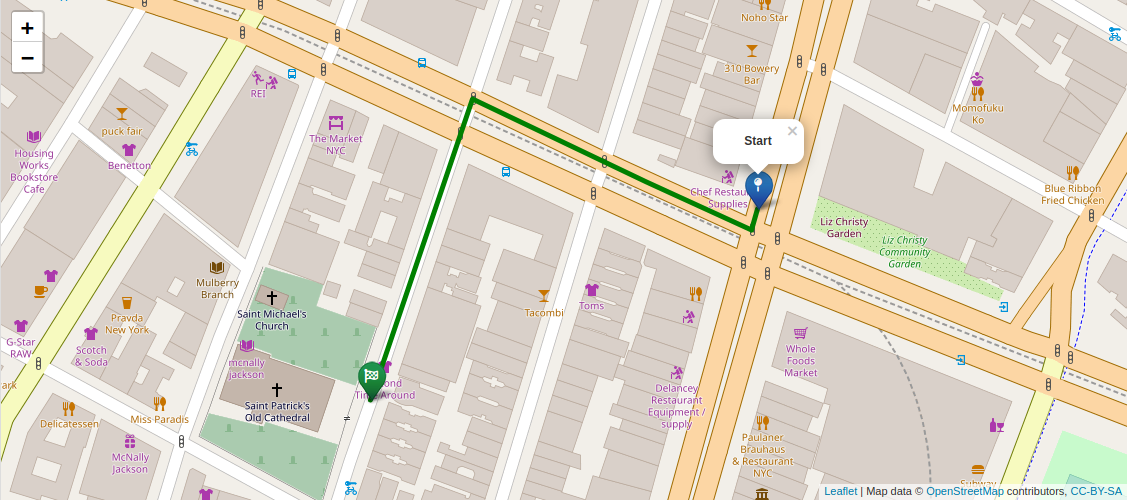}
    \scriptsize
    \textbf{reference:} \textit{Turn right at the light right ahead of you. Go through the following light. The next light is a double intersection. Turn left at the first of the lights, and go through the second one. Pass a graveyard on the right and stop at the end of it, right as the church starts. It's about 2/3 down the block.} \\
    \textbf{rule based:} \textit{Garden left light right Chef Restaurant Supplies right light straight bus stop left light left light straight grave yard right stop Saint Patrick's right}\\
    \textbf{seq2seq:} \textit{Go to the light and turn right. Go through the following light, Restaurant on the corner. Pass a bus stop on the right, and go through the light. At the following light, bus stop on the corner, turn left. Go through the following light, and stop after a step or two. Saint Mary's church will be on the right.}\\
    \textbf{graph2text:} \textit{Turn right at the first set of lights. Go through the next set of lights and then turn left at the following lights. Pass Second Time Around half way down the block. Stop at Saint's around Saint's church on the right.}\\
    \textbf{g2t+pretrain:} \textit{Go to the light and turn right. Go straight through 1 light and at the following light, there should be a bus stop on the far left corner. Turn left and go about 1/2 to the next light, stopping in front of Saint Patrick's church on the right and graveyard Memorial's on the right.}
    \caption{Route from unseen test set with successful navigation for g2t+pretrain.}
\end{figure*}

\begin{figure*}[h]
    \includegraphics[width=0.99\textwidth]{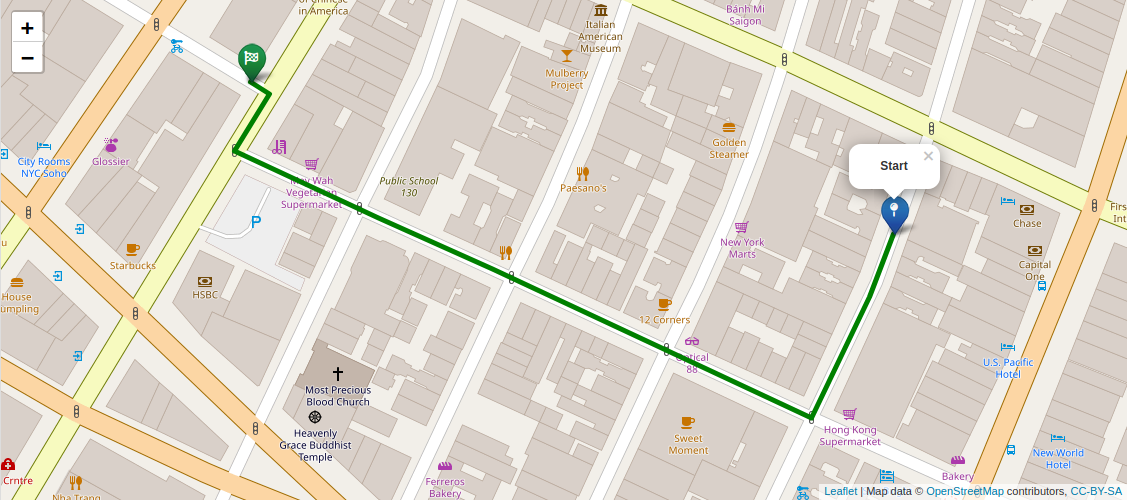}
    \scriptsize
    \textbf{reference:} \textit{Go to the light and turn right. Go through the following light with Optical 88. The next light will have Da Gennaro on the far right corner, go through the light. Go through the following light as well, with a school on the corner. Turn right at the following light. Take the first left possible and stop after a few steps.} \\
    \textbf{rule based:} \textit{Hong Kong Supermarket left light right Sweet Moment left light straight 12 Corners right light straight Da Gennaro right Public School 130 right light straight parking left May Wah Vegetarian right Hair Lounge right light right intersection left stop}\\
    \textbf{seq2seq:} \textit{Go straight and take a right at the intersection where Hong Kong supermarket is. Go through the next three intersections and at the fourth one take a right and stop at Hair Lounge.}\\
    \textbf{graph2text:} \textit{Go to the light and turn right. Go through the following light, Optical 88 on the corner. Go through the following light as well, Da Gennaro on the corner. At the following light, Hair Lounge on the corner, turn right. Take a step and stop.}\\
    \textbf{g2t+pretrain:} \textit{Head to the light and turn right. Go past the next 2 lights with Da Gennaro on the right corner. At the 3rd light with May Wah Vegetarian on the far right corner, turn left. Take one step and stop.}
    \caption{Route from partially seen test set with unsuccessful navigation for g2t+pretrain.}
\end{figure*}

\begin{figure*}[h]
    \includegraphics[width=0.99\textwidth]{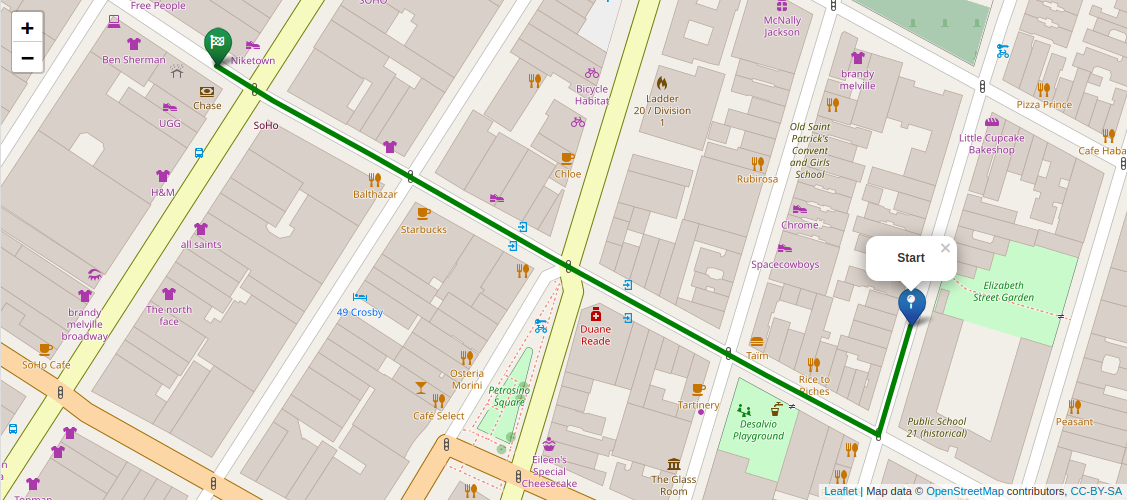}
    \scriptsize
    \textbf{reference:} \textit{Go straight to the light and turn right. Go through the next 3 lights. At the 4th, there is a Chase and Nike store on the corners. Go through the intersection and stop on the other side at your destination.} \\
    \textbf{rule based:} \textit{Public School 21 left light right Lombardi's Pizza left park left Rice to Riches right drinking water left playground left Taïm right light straight Tartinery left Duane Reade left light straight restaurant left Premium Laces right Starbucks left light straight MUJI at MoMA right Balthazar left SoHo left light straight Niketown right stop Chase left Chase left shelter left}\\
    \textbf{seq2seq:} \textit{Head to the light and turn right. Go past the next 3 lights with Starbucks on the left corner. At the 4th light with Chase bank on the near left corner, cross the street and stop.}\\
    \textbf{graph2text:} \textit{Head to the light and turn right. Go past the next 2 lights with Starbucks on the far left corner. At the 3rd light with Chase on the far left corner, cross the street and stop. Chase should be on your left.}\\
    \textbf{g2t+pretrain:} \textit{Head to the light and turn right. Go past the next 3 lights with Starbucks on the left corner. At the 4th light with Chase on the near left corner, cross the street. Take one step and stop.}
    \caption{Route from unseen test set with unsuccessful navigation for g2t+pretrain.}
\end{figure*}

\end{document}